\documentclass[conference]{IEEEtran}
\IEEEoverridecommandlockouts
\usepackage{cite}
\usepackage{amsmath,amssymb}
\usepackage{graphicx}
\usepackage{booktabs}
\usepackage{multirow}
\usepackage[hyphens]{url}
\usepackage[hidelinks]{hyperref}
\usepackage{orcidlink}
\graphicspath{{figures/}}

\begin{document}

\title{Leakage- and Label-Aware Benchmarking of Class-Imbalance Strategies for Multimodal Chest X-Ray and EHR Classification}

\author{
\IEEEauthorblockN{Dang-Khoa Bui\IEEEauthorrefmark{1}\IEEEauthorrefmark{3}~\orcidlink{0009-0007-7839-2437},
Quang-Huy Tran\IEEEauthorrefmark{1}\IEEEauthorrefmark{3}~\orcidlink{0009-0005-3172-9093},
Minh-Khoi Nguyen-Bui\IEEEauthorrefmark{1}\IEEEauthorrefmark{3}~\orcidlink{0009-0002-3419-8097},
Duc-Tuan Ngo\IEEEauthorrefmark{1}\IEEEauthorrefmark{3}~\orcidlink{0009-0005-6854-8632},\\
Thanh-Trong Tran\IEEEauthorrefmark{1}\IEEEauthorrefmark{3}~\orcidlink{0009-0004-6962-4425},
Tuan-Khoi Nguyen\IEEEauthorrefmark{1}\IEEEauthorrefmark{2}~\orcidlink{0000-0001-8556-5876},
Hoang-Anh Ngo\IEEEauthorrefmark{3}~\orcidlink{0000-0002-7583-753X}}
\IEEEauthorblockA{\IEEEauthorrefmark{1}Faculty of Computer Science and Engineering,\\ Ho Chi Minh City University of Technology (HCMUT), VNU-HCM, Vietnam}
\IEEEauthorblockA{\IEEEauthorrefmark{2}Vietnam National University Ho Chi Minh City, Vietnam}
\IEEEauthorblockA{\IEEEauthorrefmark{3}Applied AI \& Data Science Division, AK Technologies Company Limited,\\ Ho Chi Minh City, Vietnam}
% Corresponding author: hoang-anh.ngo@aktech.ai.vn
}

\maketitle

{\renewcommand{\thefootnote}{}%
 \footnotetext{Corresponding author: Dang-Khoa Bui, khoa.buidkhoa@hcmut.edu.vn}%
 \footnotetext{Senior author: Hoang-Anh Ngo, hoang-anh.ngo@aktech.ai.vn}}

\begin{abstract}
Benchmarks of class-imbalance strategies for multimodal chest X-ray (CXR) and electronic health record (EHR) prediction are often evaluated on record-level splits of cohorts in which one patient contributes several radiographs. We audit such a benchmark on 29,083 MIMIC-CXR radiographs of 10,479 patients with eight diagnosis targets. First, the record-split benchmark was misleading: a lookup table that copies the label of a development-set radiograph with the same EHR vector nearly matched the best model, and on unseen patients every model fell to the level of predicting every case positive. Second, tracing the labels back to MIMIC-IV explained why: each patient's labels are keyword matches on the diagnosis codes of a single admission, copied, with one EHR vector per patient, to all of the patient's radiographs, so about a third of the radiographs carry labels that differ from the codes of their own admission. Third, under patient-separated evaluation with admission-aligned and rule-refined labels, the initial conclusions do not hold: in the evaluated configurations no imbalance strategy raised the observed mean macro F2 over no augmentation by more than 0.01, tuned tree ensembles showed no consistent advantage over tuned logistic regression, and the released EHR variables stayed at the level of the all-positive baseline, whereas the diagnostic history available at imaging time and image embeddings exceeded it, including, with rule-refined labels, a CheXpert-trained encoder that never saw these radiographs; RAD-DINO had seen 96.6\% of them during self-supervised pretraining. Adding either encoder to the diagnostic history gave no statistically significant F2 gain; exploratory AUROC gains were significant with RAD-DINO but not with DenseNet. We provide reusable diagnostics, a leakage-aware protocol and a checklist.
\end{abstract}

\begin{IEEEkeywords}
class imbalance, data leakage, multimodal learning, chest X-ray, electronic health records, SMOTE, benchmarking, MIMIC
\end{IEEEkeywords}

\section{Introduction}
Chest radiographs and structured EHR variables are routinely combined in clinical prediction research, and public resources such as MIMIC-CXR and MIMIC-IV make this easy to do at scale~\cite{johnson2019mimiccxr,johnson2023mimiciv}. Many of the diagnostic targets in such cohorts are rare, so studies often add a class-imbalance step: random or synthetic oversampling, class weighting or balanced ensembles~\cite{chawla2002smote,lemaitre2017imbalancedlearn}. Benchmarks that cross these strategies with feature representations and classifiers are attractive because they promise practical guidance: which sampler to use, which modality carries the signal, and whether fusion helps.

Such guidance is only as good as the evaluation protocol. In MIMIC-derived cohorts one patient usually contributes several radiographs. When labels and EHR variables are attached at the patient level, a record-level split places copies of the same patient, with the same label and the same EHR vector, on both sides of the train--test boundary. Record-wise versus subject-wise evaluation is a known pitfall in clinical machine learning~\cite{saeb2017need}, and leakage of this kind is among the most common causes of irreproducible results in machine-learning-based science~\cite{kapoor2023leakage,roberts2021pitfalls}. Its consequences for imbalance benchmarks, however, are rarely quantified.

We study this problem in a case study. We first benchmarked class-imbalance strategies on a published MIMIC CXR--EHR cohort with eight binary targets~\cite{ngo2024enhancing,mottez2023repo}, whose labels come from keyword-matched ICD codes and whose EHR variables were merged per patient, using a random split over radiographs. This initial run gave clear answers: EHR-only models won every target (macro F2 0.737 vs.\ 0.457 for early fusion and 0.283 for image-only), Random Forest and Extra Trees won every target, and different oversamplers ``won'' different targets. Auditing the run against the MIMIC source tables showed that all three conclusions follow from patient leakage and from the way the labels were built. Our aim is therefore not to recommend an augmentation strategy for each disease, but to test whether such recommendations survive leakage- and label-aware evaluation, and to identify what improves performance once they do not. Such benchmarks combine three kinds of prior knowledge with different roles: domain rules that map codes to targets, prior diagnoses known at imaging time that can serve as inputs, and representations learned by image pretraining, whose data may include the evaluation images. We ask whether the apparent contributions of clinical history and visual representations survive patient-separated, provenance-aware evaluation, and whether image information adds value beyond the diagnostic knowledge available at prediction time.

Our contributions are:
\begin{itemize}
\item \textbf{Reusable diagnostics} for benchmarks on multi-record clinical cohorts, which need no retraining or only trivial baselines: a no-learning EHR lookup table, a seen/unseen-patient split of the test set, an identifier test for tabular features, a reconstruction of the labels from source codes, and a prior-diagnosis baseline.
\item \textbf{A leakage and label audit} of the case study: three leakage channels and an exact reconstruction of the released labels from MIMIC-IV codes showing that several target names do not match the codes behind them (Section~\ref{sec:audit}).
\item \textbf{A leakage-aware re-benchmark} with record, patient-grouped, patient-level and calendar-time splits; released, admission-aligned, rule-refined and report-derived (CheXpert) labels; type-aware oversampling; tuned classifiers; and richer image representations (Sections~\ref{sec:design}--\ref{sec:results}).
\item \textbf{A checklist} for multimodal imbalance benchmarks (Section~\ref{sec:discussion}).
\end{itemize}

\section{Related Work}
\textbf{Imbalance handling.} Data-level methods (random oversampling, SMOTE~\cite{chawla2002smote} and its variants) and algorithm-level methods (cost-sensitive learning, balanced bagging and boosting~\cite{chen2004brf,liu2009easyensemble}) are standard~\cite{lemaitre2017imbalancedlearn}. Large benchmarks show that oversampling rarely helps strong, well-tuned classifiers~\cite{elor2022smote}.

\textbf{Leakage and label quality in clinical ML.} Record-wise cross-validation can grossly overestimate subject-level performance~\cite{saeb2017need}. Splitting by slice instead of by patient inflated test accuracy in brain MRI~\cite{yagis2021effect}, and surveys~\cite{roberts2021pitfalls,kapoor2023leakage} list duplicated subjects across splits and features unavailable at prediction time as leading sources of over-optimism. Labels mined from codes or reports are a second weak point: the most frequent ICD codes in MIMIC-III are under-recorded by up to 35\%~\cite{searle2020silver}, and radiologist review of public chest-radiograph labels found positive predictive values well below the documented ones~\cite{oakdenrayner2020exploring}.

\textbf{CXR--EHR fusion.} Combining radiographs with EHR data improves some diagnostic tasks~\cite{jabbour2022combining,hayat2022medfuse}, and recent benchmarks stress that the benefit depends on task, cohort and fusion design~\cite{elsharief2025medmod,yin2026multimodal}. Our cohort and image representation are inherited from a transfer-learning study~\cite{ngo2024enhancing,mottez2023repo} that used TorchXRayVision models~\cite{cohen2022torchxrayvision} pretrained on CheXpert~\cite{irvin2019chexpert}.

\section{Cohort and Leakage Audit}\label{sec:audit}
\subsection{Cohort and representations}
The analysis file contains 29,083 radiographs (one per hospital admission) from 10,479 patients in MIMIC-IV/MIMIC-CXR. A patient contributes 2.78 radiographs on average (maximum 68), and 6,074 patients have more than one. Eight binary targets (pneumonia, tuberculosis, bronchitis, arthritis, fracture, lung cancer, scoliosis, lung infection) were derived from ICD diagnosis titles (Section~\ref{sec:labels}). Table~\ref{tab:cohort} shows that the effective number of positive \emph{patients} is far smaller than the number of positive radiographs; scoliosis has only 45 positive patients.

\textbf{EHR.} Seven continuous variables (age, oxygen saturation, heart rate, temperature, haemoglobin, red and white blood cell counts) and four binary variables (diabetes, HIV, smoking, cough). Oxygen saturation, heart rate and temperature are missing in 93.8\%, 75.9\% and 76.1\% of radiographs. These are general measurements rather than disease-specific predictors; we keep them because the aim is to audit the benchmark as released, and Section~\ref{sec:v5design} adds the diagnostic history.

\textbf{Image.} Each radiograph is represented by the 18 pre-sigmoid outputs of the TorchXRayVision DenseNet-121 trained on CheXpert (\texttt{densenet121-res224-chex}), one per pathology head~\cite{cohen2022torchxrayvision,irvin2019chexpert}. Re-extracting these outputs from the MIMIC-CXR-JPG images with the library's standard preprocessing reproduced every non-constant column of the analysis file up to a linear transformation (Pearson $r=1.000$ for every column over all 29,083 radiographs). The other seven columns are identically zero ($|x|<3\times10^{-38}$); they are the seven heads that CheXpert labels do not cover. The image input is therefore 11 pathology scores, not a learned latent embedding. Views are mixed: 51.0\% AP, 18.8\% PA, 28.3\% lateral and 1.9\% unrecorded, and lateral radiographs are scored by the same pathology heads as frontal ones.

\begin{table}[t]
\centering
\caption{Released labels ($N$ = 29,083 radiographs, 10,479 patients) and label audit. Own-neg.: released positives that are negative in the radiograph's own admission; later: the subset whose label came from a later admission; cons.: positive patients whose source admission has a code consistent with the rule-refined (C) definition; A, C: prevalence (\% of radiographs) and positive patients of the admission-aligned and rule-refined labels (Section~\ref{sec:labels}).}
\label{tab:cohort}
\footnotesize\setlength{\tabcolsep}{2.4pt}
\resizebox{\columnwidth}{!}{\begin{tabular}{lrrrrrrrr}
\toprule
 & \multicolumn{3}{c}{Released labels (R)} & \multicolumn{2}{c}{Positives (\%)} & Cons. & A: prev.\,\% & C: prev.\,\% \\
\cmidrule(lr){2-4}\cmidrule(lr){5-6}
Target & Radiogr. & Patients & Prev.\,\% & own-neg. & later & (\%) & (patients) & (patients) \\
\midrule
Pneumonia      & 7,980 & 4,318 & 27.44 & 33.4 & 18.4 &  89.5 & 30.84 (7,014) & 28.03 (6,434) \\
Lung cancer    & 7,787 & 2,822 & 26.78 & 13.1 &  8.4 &  14.4 & 26.83 (3,372) & 3.54 (518) \\
Fracture       & 2,078 &   918 &  7.15 & 40.2 & 24.0 &  75.6 & 7.27 (1,563) & 5.12 (1,222) \\
Bronchitis     & 1,485 &   678 &  5.11 & 40.5 & 15.4 & 100.0 & 5.74 (1,231) & 5.74 (1,231) \\
Arthritis      & 1,191 &   428 &  4.10 & 33.2 & 26.4 &  99.5 & 4.11 (700) & 4.07 (691) \\
Lung infection &   952 &   336 &  3.27 & 56.4 & 32.9 &   5.4 & 3.78 (892) & 0.39 (96) \\
Tuberculosis   &   771 &   315 &  2.65 & 43.1 & 23.3 &  11.1 & 2.43 (514) & 0.29 (64) \\
Scoliosis      &   150 &    45 &  0.52 & 48.0 & 26.7 & 100.0 & 0.45 (86) & 0.45 (86) \\
\bottomrule
\end{tabular}}
\end{table}

\subsection{Three leakage channels}
\textbf{L1 -- Patient-level duplication.} For every one of the 6,074 multi-radiograph patients, all eight labels and all 11 EHR variables are identical across that patient's radiographs, including radiographs from different admissions years apart. A record-level split therefore copies label--feature pairs across partitions.

\textbf{L2 -- The EHR vector is an identifier.} The 29,083 radiographs carry only 9,821 distinct EHR vectors. Each patient has exactly one, and only 152 vectors are shared between patients. Any model that can memorise exact feature combinations (deep trees, nearest neighbours) can identify the patient and copy the label.

\textbf{L3 -- Admission misalignment.} A patient's condition changes between admissions: acute findings such as pneumonia or a fracture resolve, whereas chronic ones such as scoliosis persist. A label coded at one admission therefore refers to that admission, not to radiographs taken months or years apart. Labels, however, come from one admission per patient and EHR values are constant per patient (Section~\ref{sec:labels}), so most radiographs carry another admission's diagnoses, some recorded later. In the preceding study's code~\cite{mottez2023repo}, laboratory and vital-sign values are also the first row of an unsorted per-admission query with no time window, so they may postdate the radiograph, and one radiograph is kept per admission without filtering the view (hence 28.3\% lateral radiographs). Patient-separated splits remove L1 and L2; for L3 we rebuild the labels per admission and add the diagnostic history known at imaging time, whereas the released laboratory and vital-sign values remain misaligned.

\subsection{Label provenance}\label{sec:labels}
We retrieved all 75,339 admissions of the cohort's patients and their 1.13 million ICD codes from MIMIC-IV v2.2; every cohort admission matched on identifier, admission date and radiograph date. One rule reproduces all eight released labels exactly in 10,477 of 10,479 patients (first record instead: 6,488; union of admissions: 6,664): a target is positive if any ICD long title of a \emph{single} admission contains a keyword (case-insensitive), and that admission is the patient's last record in the analysis file (not in date order). Keywords are the target names, except that \emph{lung cancer} matches ``malignant neoplasm'' of any site and \emph{lung infection} matches any title ending in ``infection''. Age matches that admission in 99.99\% of patients, so the EHR values were probably copied from it too; the merge itself is not in the public code.

Two problems follow (Table~\ref{tab:cohort}). \emph{Admission misalignment:} only 36.0\% of radiographs carry the labels of their own admission; 31.9\% inherit them from an earlier and 32.0\% from a later admission. Applying the same rule to each radiograph's own admission changes at least one label for 10,707 radiographs (36.8\%), and 13--56\% of the released positives per target are negative in their own admission. \emph{Keyword semantics:} matched titles include personal and family history, screening, vaccination and negated codes. Only 11.1\% of tuberculosis-positive patients have a tuberculosis diagnosis rather than a tuberculin reaction or a history code; 75.6\% of fracture positives have a current fracture code (the most frequent match is ``age-related osteoporosis without current pathological fracture''); 14.4\% of lung-cancer positives have a primary lung or bronchus malignancy; and 5.4\% of lung-infection positives have a lung or respiratory infection. Pneumonia (89.5\%), bronchitis, arthritis and scoliosis ($\geq$99.5\%) are largely consistent with the rule-refined definitions.

\subsection{Diagnostic evidence from the initial run}\label{sec:initial}
The initial run used one fixed record-level split (52/13/35\%; 15,122/3,781/10,180 radiographs), validation-selected F2 thresholds and five seeds. Its saved validation and test predictions (no retraining) show the following.

\textbf{Overlap.} 76.9\% of test radiographs belong to patients present in training or validation, and 78.6\% have an exact EHR-vector match there. \textbf{Lookup table.} Copying the label of the first development-set (training or validation) radiograph with the same EHR vector, and predicting negative otherwise, yields a macro test F2 of 0.717. The best benchmarked models reach 0.737. \textbf{Seen vs.\ unseen patients.} We split the test set by whether the patient was seen during development (2,347 unseen radiographs; thresholds re-selected on the 1,048 validation radiographs of patients absent from training). EHR-only F2 drops from 0.930 on seen to 0.322 on unseen patients, below predicting every case positive (0.335); early fusion (0.334), image-only (0.324) and validation-weighted late fusion (0.326) behave the same way (unseen AUROC at most 0.601). \textbf{Image-only equals the trivial classifier.} Its macro F2 (0.283) matches all-positive (0.282); for pneumonia and lung cancer the selected models flagged almost every radiograph (recall $\geq$0.997). \textbf{No reliable sampler winner.} For every target, all eight sampling choices (including no augmentation) have at least one EHR-only configuration within 0.01 test F2 of the validation-selected winner.

\section{Leakage-Aware Benchmark Design}\label{sec:design}
\subsection{Protocols}
Table~\ref{tab:protocols} summarises all protocols. Except for P5, all use 5-fold cross-validation with iterative multilabel stratification~\cite{sechidis2011stratification} and an inner validation set (20\% of the outer training radiographs under P1, P1-A and P1-CX and of the outer training patients otherwise) for every selection decision. \textbf{P1} assigns radiographs to folds regardless of the patient, as in the initial run. \textbf{P2} keeps radiograph-level rows but assigns folds and inner validation sets per patient, which rules out L1 and L2 across partitions. \textbf{P3} has one row per patient, the level at which the released labels exist, and uses the P2 folds, so P2/P3 comparisons are paired. Its image features are averaged over the patient's frontal (AP/PA) radiographs, or over all radiographs for the 1,544 patients without a frontal view, including radiographs taken after the label's source admission; P3 is therefore a retrospective whole-record analysis, not prediction at one time point.

\begin{table}[t]
\centering
\caption{Evaluation protocols. Pat.: patients separated across training, validation and test. Label, EHR: from the radiograph's own admission (\checkmark) or one value per patient (--); $\circ$: presumably own admission (Section~\ref{sec:labels}), possibly measured after the radiograph. P4-H and P4-C-H replace the released EHR values by age and prior diagnoses known at imaging time.}
\label{tab:protocols}
\footnotesize\setlength{\tabcolsep}{2.5pt}
\resizebox{\columnwidth}{!}{\begin{tabular}{lllccc}
\toprule
Protocol & Folds (rows) & Labels & Pat. & Label & EHR \\
\midrule
P1 & record & R & -- & -- & -- \\
P2 & patient & R & \checkmark & -- & -- \\
P3 & patient (1 row/patient) & R & \checkmark & -- & -- \\
P1-A & record & A & -- & \checkmark & -- \\
P4, P4-C & patient & A, C & \checkmark & \checkmark & -- \\
P5 & first-image year & R, A & \checkmark & A only & -- \\
P1-/P2-CX & record / patient & CX & --\,/\,\checkmark & \checkmark & -- \\
P4-H, P4-C-H & patient & A, C & \checkmark & \checkmark & history \\
P2-al & patient (aligned only) & A, C & \checkmark & \checkmark & $\circ$ \\
P2-F, P4-F, P4-C-F & patient (frontal only) & R, A, C & \checkmark & --\,/\,\checkmark & -- \\
\bottomrule
\end{tabular}}
\end{table}

\subsection{Preprocessing and type-aware oversampling}
Continuous variables are standardised and imputed with a distance-weighted 5-NN imputer that appends missingness indicators. Both are fitted on the inner training set only. Binary variables pass through. We compare \emph{no augmentation}, \emph{class weighting}, \emph{random oversampling (ROS)}, \emph{naive SMOTE} on the full matrix (as in the initial run), and \emph{type-aware SMOTE-NC}, which declares the four binary EHR variables \emph{and all missingness indicators} categorical. Image-only inputs have no discrete columns, so they use plain SMOTE. We also include two algorithm-level ensembles: Balanced Random Forest~\cite{chen2004brf} and EasyEnsemble~\cite{liu2009easyensemble}.

\subsection{Richer image representations}
To test whether the weak image result is a property of the 11 pathology scores, we add two embeddings extracted from the same 29,083 images: the 1024-dimensional pooled features of the same CheXpert DenseNet, and the 768-dimensional CLS embedding of RAD-DINO, a ViT-B/14 chest-radiograph encoder trained self-supervised with DINOv2~\cite{perezgarcia2025raddino} (public checkpoint \texttt{microsoft/rad-dino}). By image identifier, its published training list contains 28,102 (96.6\%) of our radiographs and all radiographs of 10,251 of the 10,479 patients; the list contains no MIMIC-CXR image of the other 228 patients (981 radiographs), who are too few to evaluate separately, so the CheXpert-only DenseNet serves as a control without MIMIC-CXR exposure. Both use the published checkpoints, standard preprocessing, fp16 inference and float16 storage (\texttt{scripts/01\_extract\_embeddings.py}); the DenseNet features of the history fusions were re-extracted later with the same weights and preprocessing (Section~\ref{sec:robust}). Each embedding is standardised and reduced to 64 principal components fitted on the inner training set; the number of components was fixed in advance, before any evaluation (Section~\ref{sec:robust} varies it). They enter as image-only and early-fusion representations under P2 and P3.

\subsection{Classifiers, selection and metrics}
Classifiers: logistic regression ($L_2$, $C=1$), Random Forest and Extra Trees~\cite{breiman2001randomforests,geurts2006extratrees} (300 trees, unlimited depth, minimum leaf size 1), LightGBM (300 trees, learning rate 0.05, 31 leaves) and XGBoost (300 trees, depth 6, learning rate 0.05)~\cite{ke2017lightgbm,chen2016xgboost}. Section~\ref{sec:robust} varies depth, leaf size and the number of trees. Other settings are library defaults (scikit-learn 1.6.1, imbalanced-learn 0.14.2, LightGBM 4.6.0); samplers balance the classes with $k=\min(5,n_{\min}-1)$ neighbours; folds, samplers and models use seed 42, PCA seed 0 and the inner split of fold $j$ seed $4200+j$ (Section~\ref{sec:robust} varies the model seeds). Within each outer fold, preprocessing and model are fitted on the inner training set, the threshold that maximises inner-validation F2 over $\{0.01,\dots,0.99\}$ (ties: higher recall, then precision, then closeness to 0.5) is fixed, and the model is evaluated once on the outer test fold. For each protocol, target, representation and outer fold, the configuration (strategy, classifier) with the highest F2 on that fold's inner validation set is evaluated on that fold's test set, so under the patient-separated protocols no test patient enters any choice (under P1 and P1-A, other radiographs of a test patient can be in training and validation);

We report outer-test F2, PR-AUC (average precision), AUROC and, for P1--P3, the Brier score. F2 weights recall four times as heavily as precision, reflecting screening use in which a missed case costs more than a false alarm, and was the metric of the initial run (a comparison metric, not evidence of clinical utility); since the threshold is tuned for F2 and can partly absorb the effect of weighting or oversampling, we read F2 together with the threshold-free PR-AUC and AUROC. Baselines are \emph{all-positive} ($F2=5p/(4p+1)$) and the \emph{EHR lookup table}, which copies the label of the first inner-training radiograph with the same EHR vector (fallback label chosen on inner validation; unlike in Section~\ref{sec:initial}, validation radiographs are not looked up). Fold-level results are not independent: two outer training sets share three of their four folds, so targets, classifiers and folds cannot be treated as independent samples. All intervals and tests therefore use one value per fold, the macro average over targets (and, for strategy effects, over classifiers), and the corrected resampled $t$-test~\cite{nadeau2003inference}, which replaces the factor $1/n$ in the variance of the mean by $1/n+n_{\mathrm{test}}/n_{\mathrm{train}}$ ($n=5$ folds, or 10 over two fold assignments; $n_{\mathrm{test}}/n_{\mathrm{train}}=0.2/0.64$). Margins over all-positive are Bonferroni-corrected within each table, the eight planned F2 comparisons of the encoder control (Section~\ref{sec:repaired}) are Holm-corrected together, and strategy effects (paired against no augmentation with the same classifier) are Holm-corrected~\cite{holm1979} over the 58 strategy$\times$representation$\times$protocol tests; all other p-values are unadjusted and exploratory. With five folds these tests have little power, so a non-significant difference does not establish equivalence. The grid comprises 680 tasks and 15,680 model fits, all of which completed.

\subsection{Admission-aligned labels and a calendar-time split}\label{sec:repairdesign}
Besides the released labels (R) we evaluate \emph{admission-aligned} labels (A), obtained by applying the reconstructed rule to each radiograph's own admission, and \emph{rule-refined} labels (C), which also drop history, screening, vaccination, tuberculin-reaction, ``Encounter for\dots'' and negated titles (injury titles with ``initial encounter'' are kept) and restrict lung cancer to primary lung or bronchus malignancy and lung infection to lung or respiratory infection. The task is to predict the diagnoses coded for the radiograph's admission, which need not be visible on the radiograph or present at the moment of imaging; a missing code counts as negative; all labels are rule-defined, and the rule-refined labels are not clinically verified. Laboratory and vital-sign values were not re-extracted per admission and are kept as released. Four further protocols use a reduced grid (EHR, 11 scores, RAD-DINO image and RAD-DINO fusion; no augmentation, class weighting, ROS and SMOTE/SMOTE-NC; logistic regression, LightGBM and Extra Trees): \textbf{P1-A}, the P1 folds with A labels (EHR and 11 scores); \textbf{P4} and \textbf{P4-C}, the P2 patient folds with A and C labels; and \textbf{P5}, a single calendar-time split with R and A labels. MIMIC shifts dates per patient, so ordering patients by recorded dates is close to random. We therefore estimate each radiograph's real year from the patient's anchor-year group and assign patients by their first radiograph (64/16/20\% train/validation/test; about 2010--2015 vs.\ 2015--2018), with 95\% confidence intervals from 1,000 patient-level bootstrap resamples. Later radiographs of training patients can fall in the test period (30.5\% of training radiographs), so P5 tests transfer to later-enrolled patients rather than a strict temporal cut-off. As a further baseline, a \emph{prior-diagnosis} rule predicts positive if the target was coded in any earlier admission of the patient, which is known at imaging time. These runs add 5,568 fits.

\subsection{A second label source and time-aligned history}\label{sec:v5design}
To test whether the findings depend on how the released labels were built, we also use the CheXpert labels that MIMIC-CXR-JPG provides for the report of each study~\cite{johnson2019mimiccxrjpg,irvin2019chexpert}. They are defined per radiograph rather than per patient. We take eight observations (pneumonia, lung opacity, pleural effusion, cardiomegaly, edema, pneumothorax, fracture, lung lesion; prevalence 1.4--27.9\%) and count uncertain and unmentioned findings as negative (U-zeros). Among patients with several radiographs, the label varies within the patient for 31--56\% of patients for the five common findings and 5--10\% for the three rare ones; the released labels never vary. We compare the P1 record folds with the P2 patient folds using the reduced grid above. As an analogue of the prior-diagnosis rule, a \emph{prior-study} rule predicts positive if an earlier radiograph of the patient was positive.

To test whether time-aligned EHR information helps, we build a time-aligned history vector for each radiograph: age at the radiograph's own admission and eight flags that mark whether each target was coded in an earlier admission (A or C definition), counting only earlier admissions that ended before the radiograph's admission began, so the flags are known at imaging time if an admission's codes are available once it has ended (this excludes one admission that overlapped the radiograph's admission by 15 hours and removes one pneumonia flag). It enters alone and fused with RAD-DINO on the P2 folds with A labels (\textbf{P4-H}) and C labels (\textbf{P4-C-H}). These runs add 5,760 fits. As an encoder control without MIMIC-CXR exposure, the DenseNet features enter alone and fused with the released EHR variables (P4, P4-C) and with the history (P4-H, P4-C-H), with the same folds, labels, preprocessing, grid and selection rule (2,880 fits); the history and both history fusions contain the same age variable. Table~\ref{tab:robust} lists the robustness checks with their fits (34,272; 64,160 in total).

\section{Results}\label{sec:results}
\subsection{Protocol effect}
Table~\ref{tab:protocol} summarises macro test performance. Under P1, the selected EHR-only models reach a macro F2 of 0.791 and an AUROC of 0.960. This is essentially the score of the no-learning lookup table (0.788), which finds an exact EHR match for 78.4\% of test radiographs. Once patients are separated, the lookup hit rate falls to 7.2--7.4\% (patients who happen to share an EHR vector) and its F2 to the all-positive level. EHR-only F2 falls to 0.280 under P2 ($-0.002$ relative to all-positive) and 0.320 under P3 ($+0.010$). P3 counts each patient once, which raises the all-positive level (0.310 vs.\ 0.282), so P2 and P3 are compared by their margins. Restricting P2 to the grid of P1 (same features, classifiers and strategies) changes its macro F2 by at most 0.002, so P1 and P2 differ only in the split. Under P2 and P3 no representation exceeds the all-positive baseline by more than 0.038 macro F2, and the best AUROC is 0.661.

Per target, the EHR-only model collapses most from P1 to P3 for the rare targets: scoliosis falls from 0.860 to 0.000 (all-positive: 0.021), lung infection from 0.775 to 0.136 (0.142) and tuberculosis from 0.731 to 0.139 (0.134).

\begin{table*}[t]
\centering
\caption{Outer-test performance, averaged over the 8 targets within each fold (macro) and then over the 5 folds: mean$\pm$SD; configuration selected on inner validation. $\Delta$: F2 minus all-positive F2. 95\% CIs and $^{*}$ ($p<0.05/17$, Bonferroni over the 17 rows) from the corrected resampled $t$-test over the 5 fold-level values; CIs are not adjusted for multiplicity. Brier: lower is better; predicting the training prevalence scores 0.077 (P1, P2) and 0.085 (P3). Italic: reference baselines without learning.}
\label{tab:protocol}
\footnotesize\setlength{\tabcolsep}{4pt}
\begin{tabular}{lcccccc}
\toprule
Representation & F2 & F2 95\% CI & $\Delta$ [95\% CI] & PR-AUC & AUROC & Brier \\
\midrule
\multicolumn{7}{l}{\textit{P1 record split}}\\
\quad EHR only & 0.791$\pm$0.012 & [0.766, 0.815] & +0.509$^{*}$ [+0.484, +0.533] & 0.842$\pm$0.010 & 0.960$\pm$0.005 & 0.025$\pm$0.001 \\
\quad Image: 11 scores & 0.283$\pm$0.003 & [0.278, 0.289] & +0.001 [$-$0.004, +0.007] & 0.118$\pm$0.004 & 0.569$\pm$0.007 & 0.163$\pm$0.031 \\
\quad Fusion: 11 scores & 0.560$\pm$0.002 & [0.556, 0.564] & +0.278$^{*}$ [+0.274, +0.282] & 0.540$\pm$0.016 & 0.834$\pm$0.007 & 0.062$\pm$0.002 \\
\quad \textit{All-positive} & \textit{0.282} &  &  &  & \textit{0.500} &  \\
\quad \textit{EHR lookup (hit 78.4\%)} & \textit{0.788$\pm$0.008} &  & \textit{+0.506} &  &  &  \\
\midrule
\multicolumn{7}{l}{\textit{P2 patient-grouped}}\\
\quad EHR only & 0.280$\pm$0.008 & [0.264, 0.296] & $-$0.002 [$-$0.013, +0.009] & 0.144$\pm$0.005 & 0.592$\pm$0.017 & 0.162$\pm$0.010 \\
\quad Image: 11 scores & 0.277$\pm$0.005 & [0.267, 0.286] & $-$0.005 [$-$0.017, +0.007] & 0.117$\pm$0.004 & 0.559$\pm$0.011 & 0.196$\pm$0.015 \\
\quad Fusion: 11 scores & 0.285$\pm$0.010 & [0.266, 0.305] & +0.004 [$-$0.010, +0.017] & 0.146$\pm$0.007 & 0.594$\pm$0.022 & 0.156$\pm$0.015 \\
\quad Image: DenseNet & 0.287$\pm$0.007 & [0.273, 0.301] & +0.005 [$-$0.007, +0.017] & 0.126$\pm$0.004 & 0.587$\pm$0.016 & 0.194$\pm$0.006 \\
\quad Fusion: DenseNet & 0.295$\pm$0.011 & [0.274, 0.316] & +0.013 [$-$0.002, +0.028] & 0.151$\pm$0.009 & 0.614$\pm$0.023 & 0.145$\pm$0.017 \\
\quad Image: RAD-DINO & 0.296$\pm$0.010 & [0.276, 0.316] & +0.014 [$-$0.003, +0.031] & 0.144$\pm$0.008 & 0.614$\pm$0.020 & 0.166$\pm$0.016 \\
\quad Fusion: RAD-DINO & 0.298$\pm$0.017 & [0.265, 0.332] & +0.016 [$-$0.012, +0.045] & 0.162$\pm$0.007 & 0.627$\pm$0.019 & 0.157$\pm$0.024 \\
\quad \textit{All-positive} & \textit{0.282} &  &  &  & \textit{0.500} &  \\
\quad \textit{EHR lookup (hit 7.2\%)} & \textit{0.286$\pm$0.004} &  & \textit{+0.004} &  &  &  \\
\midrule
\multicolumn{7}{l}{\textit{P3 patient-level}}\\
\quad EHR only & 0.320$\pm$0.005 & [0.310, 0.331] & +0.010 [+0.000, +0.020] & 0.175$\pm$0.004 & 0.607$\pm$0.011 & 0.167$\pm$0.019 \\
\quad Image: 11 scores & 0.306$\pm$0.012 & [0.283, 0.330] & $-$0.004 [$-$0.027, +0.019] & 0.150$\pm$0.003 & 0.570$\pm$0.013 & 0.183$\pm$0.030 \\
\quad Fusion: 11 scores & 0.320$\pm$0.010 & [0.301, 0.339] & +0.010 [$-$0.009, +0.028] & 0.183$\pm$0.003 & 0.621$\pm$0.013 & 0.163$\pm$0.012 \\
\quad Image: DenseNet & 0.320$\pm$0.011 & [0.297, 0.342] & +0.010 [$-$0.014, +0.033] & 0.171$\pm$0.008 & 0.609$\pm$0.017 & 0.144$\pm$0.031 \\
\quad Fusion: DenseNet & 0.322$\pm$0.017 & [0.288, 0.356] & +0.012 [$-$0.023, +0.046] & 0.194$\pm$0.002 & 0.634$\pm$0.019 & 0.127$\pm$0.020 \\
\quad Image: RAD-DINO & 0.323$\pm$0.012 & [0.299, 0.347] & +0.013 [$-$0.011, +0.036] & 0.193$\pm$0.006 & 0.639$\pm$0.011 & 0.143$\pm$0.019 \\
\quad Fusion: RAD-DINO & 0.348$\pm$0.016 & [0.316, 0.380] & +0.038 [+0.007, +0.070] & 0.217$\pm$0.014 & 0.661$\pm$0.018 & 0.135$\pm$0.027 \\
\quad \textit{All-positive} & \textit{0.310} &  &  &  & \textit{0.500} &  \\
\quad \textit{EHR lookup (hit 7.4\%)} & \textit{0.313$\pm$0.001} &  & \textit{+0.002} &  &  &  \\
\bottomrule
\end{tabular}
\end{table*}

\textbf{Seen vs.\ unseen patients within P1.} Inside P1 (4,715 seen and 1,101 unseen test radiographs per fold), EHR-only F2 is $0.942\pm0.003$ on seen and $0.227\pm0.014$ on unseen patients, or $0.336\pm0.006$ with the threshold re-selected on unseen validation patients, the all-positive level ($0.335$); early fusion and image-only models behave the same way.

\textbf{Which classifier wins depends on the split.} With EHR inputs and no augmentation, Extra Trees (0.793), Random Forest (0.788) and LightGBM (0.742) dominate under P1, and logistic regression trails at 0.313. Under P2 and P3 the order inverts: logistic regression is best (0.297 and 0.326) and the tree ensembles are worst (0.278--0.281 and 0.308--0.314).

\subsection{Imbalance strategies under patient-separated evaluation}
\begin{figure}[t]
\centering
\includegraphics[width=\columnwidth]{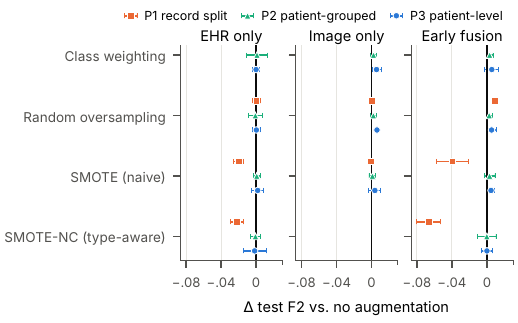}
\caption{Paired change in test F2 vs.\ no augmentation (same classifier, target and fold) for EHR, image (11 scores, DenseNet, RAD-DINO) and fusion inputs. Points: mean of five fold-level differences averaged over targets, classifiers and representations; bars: 95\% CI (corrected resampled $t$-test).}
\label{fig:strategies}
\end{figure}

Fig.~\ref{fig:strategies} shows paired changes in test F2 relative to no augmentation with the same classifier, target and fold, separately for EHR, image and fusion inputs. Under P2 and P3 the mean changes lie between $-0.002$ and $+0.006$, with upper 95\% limits of at most 0.014; the few intervals that exclude zero (image and fusion inputs under P3) correspond to gains below 0.007. Of the 58 strategy$\times$representation$\times$protocol tests, two remain significant after Holm correction, both losses under P1; no gain under P2 or P3 (at most 0.009) is significant. When the best classifier is selected separately for each strategy, no augmentation comes within 0.007 macro F2 of the best strategy for every representation under P2 and P3. Balanced Random Forest and EasyEnsemble reach the same level (P3 RAD-DINO fusion: 0.344 and 0.334 vs.\ 0.344). Type-aware SMOTE-NC, which keeps binary and missingness columns binary (naive SMOTE gave them non-binary values in 11.2--77.4\% of synthetic rows), is never better than naive SMOTE and lowers PR-AUC slightly ($-0.001$ to $-0.016$). Rebalancing degrades the probabilities: relative to no augmentation it raises the Brier score by 0.02--0.08 under P2 and P3 (mean over classifiers; logistic regression alone: 0.10--0.17), and because selection optimises F2, most selected models are rebalanced and score far worse than predicting the training prevalence (Table~\ref{tab:protocol}), consistent with~\cite{vandengoorbergh2022harm}.

Under P1, interpolating samplers \emph{lower} F2 (SMOTE $-0.020$, 95\% CI $[-0.027,-0.012]$; SMOTE-NC $-0.044$ $[-0.054,-0.035]$), whereas random oversampling, which only duplicates rows, does not.

\subsection{Modality comparison}
Under patient-separated evaluation the modality ordering of the initial run reverses (Table~\ref{tab:protocol}), but with the released labels the differences are small. Under P3, the inherited 11 pathology scores are at the trivial level (0.306; AUROC 0.570); RAD-DINO raises image-only AUROC to 0.639 ($+0.069$, $p=0.009$) and F2 to 0.323 ($+0.016$, $p=0.10$). RAD-DINO matches the released EHR variables on F2 (0.323 vs.\ 0.320, $p=0.81$) and has a higher AUROC (0.639 vs.\ 0.607, $p=0.10$). Early fusion with RAD-DINO gives the best result (0.348; AUROC 0.661; $+0.028$ F2 and $+0.054$ AUROC over EHR, $p=0.11$ and $p=0.04$). Under P2, RAD-DINO exceeds EHR by 0.016 F2 ($p=0.02$). All representations nevertheless discriminate weakly with the released labels (AUROC $\le$ 0.75 for every target).

\subsection{Admission-aligned and rule-refined labels}\label{sec:repaired}
Table~\ref{tab:repaired} repeats the analysis with admission-aligned and rule-refined labels.

\textbf{Most of the record-split advantage came from the copied labels.} Under P1-A, the EHR-only model falls from 0.791 to 0.421 macro F2, and its margin over the all-positive baseline from 0.509 to 0.131. The lookup table still finds an exact EHR match for 78.4\% of test radiographs but now reaches only 0.355, because a patient's other admissions no longer share the label. The remaining margin is still leakage: seen patients score 0.449 and unseen ones 0.289 (all-positive 0.336). It reflects diagnoses that persist across admissions; scoliosis, for instance, still scores 0.297 against an all-positive level of 0.022.

\textbf{With patients separated, the image carries the signal.} Under P4 the released EHR variables and the 11 pathology scores stay at the all-positive level. The weak EHR result also persists in the source-admission subset (Section~\ref{sec:robust}), although re-timed laboratory and vital-sign values were not tested. RAD-DINO image and fusion models, by contrast, gain 0.025 and 0.027 (AUROC 0.637 and 0.650; RAD-DINO vs.\ EHR $+0.022$ F2 and $+0.058$ AUROC, $p=0.02$ and $p=0.003$). With rule-refined labels (P4-C) the gap widens: RAD-DINO gains 0.054 over all-positive and exceeds EHR by 0.047 F2 and 0.099 AUROC ($p=0.008$ and $p=0.005$), and adding EHR to RAD-DINO gives no significant gain ($+0.005$, $p=0.60$). After Holm adjustment (Table~\ref{tab:encoder}), the CheXpert-only DenseNet exceeds all-positive with C labels but not with A labels, and RAD-DINO's advantage over it (0.019 and 0.032 F2) is not significant. The control supports image signal without MIMIC-CXR exposure; how much the overlap adds to RAD-DINO is undetermined, because the encoders also differ in architecture, objective and training data. Lung cancer is the clearest case: for the admission-aligned label built with the released keyword rule (any malignancy, including history codes) RAD-DINO reaches an AUROC of 0.650 under P4, and for primary lung malignancy (P4-C) 0.818 (F2 0.369 vs.\ 0.155 all-positive); the rule-refined label changes the target and its prevalence as well as the rule. The calendar-time split gives the same ordering (Table~\ref{tab:repaired}).

\textbf{Diagnostic history is the strongest input.} Predicting positive whenever the target was coded in an earlier admission reaches 0.385 macro F2 under P4 and 0.337 under P4-C, above every model trained on the released features (best 0.318 and 0.245). The rule is strongest for chronic conditions (under P4: arthritis 0.525, lung cancer 0.741, scoliosis 0.385) and weak for pneumonia (0.272 vs.\ 0.690 all-positive); none of the 11 released EHR variables encodes this history. Models on the time-aligned history vector reach 0.439 (P4-H) and 0.371 (P4-C-H), $+0.145$ and $+0.177$ F2 over the released EHR variables ($p<0.001$; the gain comes from the prior-diagnosis flags, Section~\ref{sec:robust}), and beat the rule by $+0.055$ ($p<0.001$) with admission-aligned labels; with rule-refined labels the margin ($+0.033$, $p=0.06$) is not significant. Fusing history with RAD-DINO gives the highest macro AUROC among the patient-separated A/C models in Table~\ref{tab:repaired} (0.766 and 0.762; $+0.025$ and $+0.049$ over history alone, unadjusted $p\le0.010$), while F2 changes little ($+0.005$ and $+0.013$, not significant). With re-extracted DenseNet features (Section~\ref{sec:robust}), the history fusion shows no statistically significant improvement over history alone in F2 (Table~\ref{tab:encoder}) or AUROC ($+0.009$ and $+0.016$, $p=0.29$ and $0.08$), and its AUROC is 0.016 and 0.033 below that of the RAD-DINO history fusion ($p=0.07$ and $0.02$). Without MIMIC-CXR exposure, these results therefore do not establish that the image adds to the time-aligned history; whether the AUROC gain with RAD-DINO reflects a stronger encoder or its pretraining overlap is undetermined.

\textbf{Strategies and classifiers.} Under P4 and P4-C every paired mean change relative to no augmentation lies between $-0.010$ and $+0.004$ F2; none is significant after Holm correction. Under P1-A, SMOTE-NC again lowers F2 ($-0.021$). With history features, every rebalancing strategy lowers the F2 of the history-only model ($-0.015$ to $-0.010$), and SMOTE-NC lowers history fusion by 0.027 (Holm-significant with A labels).

\begin{table*}[t]
\centering
\caption{Admission-aligned (A) and rule-refined (C) labels: macro outer-test performance (8 targets). P1-A, P4, P4-C: mean$\pm$SD over 5 folds; 95\% CIs and $^{*}$ ($p<0.05/14$, Bonferroni over the 14 unmarked model rows) from the corrected resampled $t$-test over the fold-level values; $^{\ddagger}$ CheXpert-only DenseNet control, tested in its own Holm-corrected family (Table~\ref{tab:encoder}), not with $^{*}$; image and EHR-fusion rows use the original features, history-fusion rows the re-extracted ones (Section~\ref{sec:robust}). P5: single calendar-time split, 95\% CIs from 1,000 patient-level bootstrap resamples (no SD); $^{\dagger}$ bootstrap CI of $\Delta$ excludes 0. Italic: baselines without learning.}
\label{tab:repaired}
\footnotesize\setlength{\tabcolsep}{5pt}
\begin{tabular}{lccccc}
\toprule
Representation & F2 & F2 95\% CI & $\Delta$ [95\% CI] & PR-AUC & AUROC \\
\midrule
\multicolumn{6}{l}{\textit{P1-A record split, admission-aligned labels}}\\
\quad EHR only & 0.421$\pm$0.013 & [0.395, 0.448] & +0.131$^{*}$ [+0.112, +0.149] & 0.287$\pm$0.015 & 0.719$\pm$0.012 \\
\quad Image: 11 scores & 0.290$\pm$0.003 & [0.285, 0.296] & $-$0.000 [$-$0.010, +0.009] & 0.132$\pm$0.003 & 0.567$\pm$0.010 \\
\quad \textit{All-positive / lookup (hit 78.4\%)} & \textit{0.291 / 0.355} &  &  &  &  \\
\midrule
\multicolumn{6}{l}{\textit{P4 patient-grouped, admission-aligned labels}}\\
\quad EHR only & 0.294$\pm$0.007 & [0.281, 0.308] & +0.004 [$-$0.008, +0.015] & 0.140$\pm$0.005 & 0.579$\pm$0.008 \\
\quad Image: 11 scores & 0.289$\pm$0.002 & [0.285, 0.293] & $-$0.002 [$-$0.006, +0.003] & 0.131$\pm$0.004 & 0.566$\pm$0.016 \\
\quad Image: DenseNet$^{\ddagger}$ & 0.297$\pm$0.006 & [0.285, 0.309] & +0.007 [$-$0.000, +0.013] & 0.143$\pm$0.005 & 0.592$\pm$0.010 \\
\quad Fusion: DenseNet$^{\ddagger}$ & 0.304$\pm$0.009 & [0.287, 0.321] & +0.013 [+0.000, +0.027] & 0.160$\pm$0.006 & 0.614$\pm$0.014 \\
\quad Image: RAD-DINO & 0.316$\pm$0.004 & [0.307, 0.325] & +0.025$^{*}$ [+0.015, +0.036] & 0.173$\pm$0.006 & 0.637$\pm$0.011 \\
\quad Fusion: RAD-DINO & 0.318$\pm$0.008 & [0.302, 0.334] & +0.027$^{*}$ [+0.018, +0.037] & 0.180$\pm$0.004 & 0.650$\pm$0.010 \\
\quad History (P4-H) & 0.439$\pm$0.014 & [0.412, 0.467] & +0.149$^{*}$ [+0.124, +0.173] & 0.260$\pm$0.014 & 0.741$\pm$0.003 \\
\quad Fusion: DenseNet + history$^{\ddagger}$ & 0.437$\pm$0.010 & [0.418, 0.456] & +0.147 [+0.129, +0.164] & 0.277$\pm$0.010 & 0.750$\pm$0.008 \\
\quad Fusion: RAD-DINO + history & 0.444$\pm$0.010 & [0.424, 0.464] & +0.154$^{*}$ [+0.134, +0.173] & 0.295$\pm$0.007 & 0.766$\pm$0.003 \\
\quad \textit{All-positive / prior diagnosis} & \textit{0.291 / 0.385} &  &  &  &  \\
\midrule
\multicolumn{6}{l}{\textit{P4-C patient-grouped, rule-refined labels}}\\
\quad EHR only & 0.193$\pm$0.009 & [0.175, 0.212] & +0.007 [$-$0.010, +0.024] & 0.092$\pm$0.005 & 0.578$\pm$0.012 \\
\quad Image: 11 scores & 0.194$\pm$0.004 & [0.185, 0.203] & +0.007 [+0.001, +0.014] & 0.088$\pm$0.002 & 0.586$\pm$0.022 \\
\quad Image: DenseNet$^{\ddagger}$ & 0.209$\pm$0.005 & [0.198, 0.219] & +0.022 [+0.012, +0.032] & 0.105$\pm$0.003 & 0.607$\pm$0.017 \\
\quad Fusion: DenseNet$^{\ddagger}$ & 0.212$\pm$0.005 & [0.202, 0.223] & +0.026 [+0.020, +0.031] & 0.117$\pm$0.003 & 0.620$\pm$0.024 \\
\quad Image: RAD-DINO & 0.240$\pm$0.011 & [0.218, 0.262] & +0.054$^{*}$ [+0.031, +0.077] & 0.136$\pm$0.003 & 0.677$\pm$0.016 \\
\quad Fusion: RAD-DINO & 0.245$\pm$0.007 & [0.232, 0.258] & +0.058$^{*}$ [+0.050, +0.067] & 0.143$\pm$0.005 & 0.683$\pm$0.015 \\
\quad History (P4-C-H) & 0.371$\pm$0.018 & [0.334, 0.407] & +0.184$^{*}$ [+0.148, +0.220] & 0.206$\pm$0.024 & 0.713$\pm$0.015 \\
\quad Fusion: DenseNet + history$^{\ddagger}$ & 0.361$\pm$0.013 & [0.335, 0.387] & +0.174 [+0.154, +0.194] & 0.232$\pm$0.010 & 0.730$\pm$0.020 \\
\quad Fusion: RAD-DINO + history & 0.384$\pm$0.021 & [0.343, 0.425] & +0.197$^{*}$ [+0.157, +0.238] & 0.252$\pm$0.021 & 0.762$\pm$0.015 \\
\quad \textit{All-positive / prior diagnosis} & \textit{0.186 / 0.337} &  &  &  &  \\
\midrule
\multicolumn{6}{l}{\textit{P5 calendar-time split}}\\
\quad EHR only, R & 0.306 & [0.288, 0.322] & +0.005 [$-$0.011, +0.019] & 0.160 & 0.576 \\
\quad Fusion: RAD-DINO, R & 0.320 & [0.304, 0.333] & +0.018$^{\dagger}$ [+0.007, +0.030] & 0.180 & 0.627 \\
\quad EHR only, A & 0.315 & [0.301, 0.328] & +0.007 [$-$0.004, +0.018] & 0.161 & 0.598 \\
\quad Fusion: RAD-DINO, A & 0.336 & [0.323, 0.348] & +0.028$^{\dagger}$ [+0.019, +0.038] & 0.198 & 0.663 \\
\quad \textit{All-positive, R / A} & \textit{0.301 / 0.308} &  &  &  &  \\
\bottomrule
\end{tabular}
\end{table*}

\begin{table}[t]
\centering
\caption{Encoder control: the eight planned F2 comparisons (macro outer-test F2 over 5 folds, corrected resampled $t$-test); $p_{\mathrm{H}}$: Holm-adjusted within this family.}
\label{tab:encoder}
\footnotesize\setlength{\tabcolsep}{3pt}
\resizebox{\columnwidth}{!}{\begin{tabular}{llccc}
\toprule
Contrast & Labels & $\Delta$F2 [95\% CI] & $p$ & $p_{\mathrm{H}}$ \\
\midrule
DenseNet image $-$ all-positive & A & +0.007 [$-$0.000, +0.013] & 0.052 & 0.193 \\
 & C & +0.022 [+0.012, +0.032] & 0.004 & 0.025 \\
DenseNet + EHR $-$ all-positive & A & +0.013 [+0.000, +0.027] & 0.048 & 0.193 \\
 & C & +0.026 [+0.020, +0.031] & $<$0.001 & 0.001 \\
RAD-DINO image $-$ DenseNet image & A & +0.019 [+0.002, +0.035] & 0.034 & 0.180 \\
 & C & +0.032 [+0.005, +0.058] & 0.030 & 0.180 \\
DenseNet + history $-$ history & A & $-$0.002 [$-$0.022, +0.018] & 0.809 & 1.000 \\
 & C & $-$0.010 [$-$0.050, +0.030] & 0.534 & 1.000 \\
\bottomrule
\end{tabular}}
\end{table}

\subsection{A second label source}\label{sec:chexpert}
We repeat the protocol comparison with CheXpert report labels of the same radiographs (all-positive F2 0.370; prior-study rule 0.253).

\textbf{With per-radiograph labels the record split barely inflates.} The record split raises EHR-only F2 by only 0.021 over the patient split (0.394 vs.\ 0.373), and the margins of the image models over all-positive differ by at most 0.002 between the two splits. The lookup table still finds an exact EHR match for 78.4\% of test radiographs, but reaches only 0.297, below the all-positive level (0.370), because a patient's radiographs no longer share a label. This is consistent with patient overlap inflating results mainly when labels are shared across a patient's records, as the released labels were; the CheXpert targets also differ in definition and prevalence, so the comparison does not isolate the effect of copied labels.

\textbf{The image carries the signal.} Under P2, RAD-DINO (0.438, AUROC 0.722) exceeds the released EHR variables (0.373) by 0.065 F2 and 0.142 AUROC and the 11 pathology scores by 0.034 F2 and 0.057 AUROC ($p\le0.007$); adding EHR gives no significant gain ($+0.002$, $p=0.72$). Imbalance strategies again improve F2 by at most $0.005$ with no Holm-significant effect, SMOTE-NC lowers fusion F2 by 0.014, and logistic regression is selected in 116 of 160 fold-level choices.

\subsection{Robustness checks}\label{sec:robust}
\begin{table*}[t]
\centering
\caption{Robustness checks (Section~\ref{sec:robust}), patient-separated protocols unless stated. F2: macro outer-test F2; unadjusted p-values from the corrected resampled $t$-test. Fits: model fits of each check (34,272 in total); full grids are listed in the configuration file of the code release.}
\label{tab:robust}
\scriptsize\setlength{\tabcolsep}{2pt}
\begin{tabular}{@{}p{0.085\textwidth}p{0.85\textwidth}r@{}}
\toprule
Check & Setting, result and conclusion & Fits \\
\midrule
Tuned models & \textit{Depth 3 or 5, minimum leaf 20 or 50, 100--1,000 trees or rounds; LR $C\in\{1,0.1,0.01\}$; no augmentation.} P1: depth 3 removes the tree advantage (EHR F2: RF 0.788$\to$0.366, ET 0.793$\to$0.315; LR 0.313), yet validation picks unlimited depth in 75 of 80 choices. Tuned trees vs.\ tuned LR: $-0.017$ to $+0.005$; 1,000 boosting rounds overfit (down to $-0.057$). \textbf{No consistent tree advantage.} & 10,560 \\
\addlinespace[1pt]
Joint tuning & \textit{P4-C; for each strategy, LR ($C\in\{1,0.1,0.01\}$) or LightGBM (four settings) chosen on inner validation.} vs.\ no augmentation: EHR $+0.006$ to $+0.007$, fusion $-0.009$ to $+0.001$ (all $p\ge0.24$; CIs up to $+0.023$); tuned LightGBM vs.\ tuned LR $-0.020$ to $-0.003$. \textbf{No significant or consistent strategy gain within this grid.} & 2,240 \\
\addlinespace[1pt]
Aligned subset & \textit{The 10,479 radiographs whose own admission supplied the labels (and presumably the EHR values); laboratory and vital-sign values may postdate the radiograph, so this evaluates admission alignment, not prediction-time validity.} EHR $+0.017$ over all-positive; RAD-DINO fusion vs.\ EHR $+0.009$ ($p=0.24$); C labels: RAD-DINO vs.\ EHR $+0.028$ ($p=0.004$). \textbf{EHR stays weak.} & 4,080 \\
\addlinespace[1pt]
Frontal only & \textit{AP/PA radiographs only (20,294).} F2 vs.\ all views $-0.001$ to $+0.015$; EHR at the all-positive level; RAD-DINO vs.\ EHR $+0.013$ (R, $p=0.36$), $+0.029$ (A, $p=0.017$), $+0.050$ (C, $p=0.011$). \textbf{No conclusion changes.} & 5,760 \\
\addlinespace[1pt]
PCA size & \textit{32, 128 or 768 instead of 64 RAD-DINO components (37\% of the variance).} 128: within 0.008; 32 or 768: down to $-0.018$. \textbf{64 components adequate for RAD-DINO with LR.} & 720 \\
\addlinespace[1pt]
Age, history & \textit{Own-admission age; prior-diagnosis flags alone.} Age: within 0.001 of the released age, at the all-positive level; flags: $+0.136$ (A), $+0.180$ (C) over the released EHR. \textbf{The history gain comes from the diagnoses.} & 240 \\
\addlinespace[1pt]
Fold seed 43 & \textit{Second patient-fold assignment (fold seed 43) and its seed-42 counterpart; LR (P2: four strategies; P4, P4-C: no augmentation).} Changes of at most 0.005 per configuration; over both seeds RAD-DINO image and fusion vs.\ EHR $+0.021$, $+0.023$ (A) and $+0.049$, $+0.052$ (C); history $+0.146$, $+0.172$ ($p<0.001$). \textbf{No conclusion changes.} & 1,360 \\
\addlinespace[1pt]
Model seeds & \textit{Model seeds 1, 2, 3 and 42 (samplers, classifiers, PCA) on the seed-42 folds: P4-C (EHR, RAD-DINO image and fusion), P4-C-H (history alone; the history fusions were not repeated) and P5-A, admission grid.} Paired differences, mean (SD) over the four seeds: strategies vs.\ none within $\pm0.004$ (SD $\le0.001$, no seed $p<0.05$); RAD-DINO image and fusion vs.\ EHR $+0.045$ (0.002) and $+0.047$ (0.003), $p<0.05$ for every seed; fusion vs.\ image $+0.002$ (0.002); history vs.\ prior rule $+0.033$ (0.003), $p=0.03$--$0.06$; P5-A fusion vs.\ EHR $+0.019$ (0.003). \textbf{Effect estimates remain similar across seeds; history vs.\ prior rule crosses the nominal 0.05 threshold.} & 6,624 \\
\addlinespace[1pt]
Repeated runs & \textit{Default-seed fits rerun with the released scripts and re-extracted features: P4-C and P5-A (EHR, RAD-DINO); P4 and P4-C (DenseNet image, fitted with NumPy 2.4.6 instead of 2.1.3; versions, devices and checksums in \texttt{EMBEDDINGS.md}, unknown for the original extraction).} RAD-DINO grid: all 1,728 fits identical. DenseNet: macro F2 within 0.001 and AUROC within 0.004 of the published values, single fits up to 0.057 apart. \textbf{Fits repeat exactly for identical features; the DenseNet re-extraction is close but not identical.} & 2,688 \\
\addlinespace[1pt]
Operating points & \textit{Selected models at their inner-validation F2 threshold (P4-C, macro over targets; prevalence 6.0\%).} RAD-DINO fusion: recall 0.41, precision 0.11, specificity 0.78; EHR: 0.43, 0.07, 0.66; lung cancer (fusion): 0.47, 0.21, 0.93; pneumonia: 0.93, 0.33, 0.26; P4-C-H history: 0.50, 0.27, 0.78. \textbf{Precision too low to establish screening use.} & -- \\
\addlinespace[1pt]
Rare targets & \textit{Without scoliosis, tuberculosis and lung infection.} Strategies at most $+0.006$; released EHR within 0.021 of all-positive; RAD-DINO up to $+0.087$, history $+0.12$ to $+0.21$. \textbf{No conclusion changes.} & -- \\
\bottomrule
\end{tabular}
\end{table*}

None of the checks in Table~\ref{tab:robust} changes an effect estimate materially; across model seeds, the history-versus-prior-rule comparison with C labels crosses the nominal 0.05 threshold ($p=0.03$--$0.06$). Repeated runs test determinism, model seeds give the spread of the paired differences, and the second fold assignment varies the folds. Patient separation exposes the effective sample size: positive patients per test fold are 9 for scoliosis with released labels (15--21 with A or C labels) and, with C labels, 7--17 for tuberculosis and 12--23 for lung infection, so per-target F2 for these targets is highly uncertain.

\section{Discussion}\label{sec:discussion}
\textbf{What the leakage explains.} Three conclusions of the initial run change once patients are separated. (i) \emph{EHR over imaging:} the EHR vector identifies the patient, so EHR-only models won by recall of identity, not by clinical signal. (ii) \emph{Tree ensembles everywhere:} deep trees memorise exact feature combinations; limiting their depth removes the record-split advantage, and leaky validation still selects the deepest trees. (iii) \emph{Sampler winners:} once identity drives the score, sampler differences shrink to within 0.01 F2, and no reliable sampler winner emerges under the evaluated protocols. The copied labels amplify all three effects: with admission-aligned labels three quarters of the EHR-only margin under the record split disappears.

\textbf{Image signal and complementarity.} The evidence that radiographs carry signal is stronger than the evidence that the modalities complement each other: image-only models exceed all-positive under P4 and P4-C, including the DenseNet control with C labels, but adding EHR to RAD-DINO changes P4-C F2 only from 0.240 to 0.245 ($p=0.60$), and fusion gains are clearer in AUROC than in F2 (P3 fusion vs.\ EHR: AUROC $+0.054$, $p=0.04$; F2 $+0.028$, $p=0.11$; likewise for history plus RAD-DINO, whereas history plus DenseNet improved neither metric significantly, Section~\ref{sec:repaired}). We read the fusion results as exploratory.

\textbf{Implications for imbalance research.} Oversampling duplicates or interpolates minority patients that also appear in the test set, so leaky protocols reward strategies by how they interact with memorised minority identities. Once leakage is removed, and within the evaluated strategies, classifiers, tuning ranges and metrics, the information given to the model mattered more than the imbalance strategy: on the same targets, diagnostic history and image embeddings raised F2 far more than any sampler. Changing the label definition changed the measured relationships between modalities even more, but it also changes the target and its prevalence, so it is not a like-for-like gain.
\textbf{Checklist.} Most items are established good practice~\cite{saeb2017need,kapoor2023leakage}; we list them because each would have caught a failure in our case study:
(1) split by patient, and evaluate at the level at which labels are defined;
(2) report the number of positive \emph{patients} per target and per fold;
(3) include an all-positive baseline and a no-learning lookup baseline;
(4) for every selected model, report seen- vs.\ unseen-patient performance whenever a record split is used;
(5) select thresholds, samplers and fusion weights on inner validation data only, and declare binary variables and missingness indicators as categorical when oversampling;
(6) rebuild every label from the source codes, check which admission it comes from and what the matched codes mean, and document when each feature becomes available relative to the radiograph;
(7) when diagnostic targets can recur, include a prior-diagnosis baseline and give models the diagnostic history available at imaging time.

\textbf{Limitations.} The study uses one cohort from one hospital system. Laboratory values and vital signs were not re-extracted per admission or restricted to measurements before the radiograph; only age and diagnostic history were time-aligned. RAD-DINO saw 96.6\% of our radiographs, without labels, during pretraining~\cite{perezgarcia2025raddino}; patient separation downstream does not remove this overlap, so its results may be optimistic; how much the overlap adds is undetermined, because the DenseNet control also differs in architecture, objective and training data, and no matched encoder trained without these patients was available. Image encoders were used frozen, so imaging results may underestimate fine-tuned encoders. CheXpert labels are extracted automatically from reports and are noisy, and we used only the U-zeros policy; the rule-refined definitions are our own. The calendar-time split relies on 3-year anchor-year groups and a single split. Model seeds were varied only for P4-C, history alone under P4-C-H and P5-A, not for the history fusions, whose small differences from history alone rest on one model seed; the fold assignment was varied only for logistic regression, and strategies were combined with tuned classifiers only under P4-C (LR and LightGBM; EHR and RAD-DINO fusion). Thresholds come from a fixed grid, and calibration is reported through Brier scores only.

\section{Conclusion}
A lookup table without learning nearly reproduced the answers of a record-split imbalance benchmark whose labels were copied across each patient's radiographs. Evaluated by patient and with labels re-derived per admission, no evaluated imbalance strategy raised the observed mean macro F2 by more than 0.01, tuned tree ensembles showed no consistent advantage over tuned logistic regression, and time-aligned diagnostic history and image embeddings (with rule-refined labels, also one without MIMIC-CXR pretraining), not the released EHR variables, carried the signal; added to the history, neither encoder significantly improved F2, and an exploratory AUROC gain was significant only with RAD-DINO, which had seen these radiographs during pretraining. These conclusions are limited to the evaluated tasks, models, tuning ranges and metrics, and the operating points do not establish screening value; the contribution is the audit and the evaluation method. Imbalance benchmarks on multi-record clinical cohorts should be judged first by their split, their labels and their baselines, and only then by their rankings.

\textbf{Code availability.} Code for every step (release v1.0.0), including the embedding extraction, its configuration and an image-feature manifest, is available at \url{https://github.com/buidangkhoa-vn/Benchmarking-of-Class-Imbalance-Strategies-for-Multimodal-Chest-X-Ray-and-EHR-Classification}; MIMIC-derived data are not redistributed.

\section*{Acknowledgment}
We acknowledge Ho Chi Minh City University of Technology (HCMUT), VNU-HCM for supporting this study. This study used the MIMIC-IV and MIMIC-CXR-JPG databases, accessed under PhysioNet credentialed data use agreements; all analyses were performed on de-identified data. Claude (Anthropic) was used to refine the code and to assist in drafting the text.

\bibliographystyle{IEEEtran}
\bibliography{references}

\end{document}